\documentclass[letterpaper]{article} 
\usepackage{aaai25}  
\usepackage{times}  
\usepackage{helvet}  
\usepackage{courier}  
\usepackage[hyphens]{url}  
\usepackage{graphicx} 
\usepackage{natbib}  
\usepackage{caption} 
\usepackage{algorithm}
\usepackage{algorithmic}

\usepackage{newfloat}
\usepackage{listings}
\DeclareCaptionStyle{ruled}{labelfont=normalfont,labelsep=colon,strut=off} 
\floatstyle{ruled}
\newfloat{listing}{tb}{lst}{}
\floatname{listing}{Listing}
\usepackage{amsmath}
\usepackage{multirow}
\usepackage{booktabs}
\usepackage{subfig}
\usepackage{subcaption}
\usepackage{comment}

\title{Revisiting Video Quality Assessment from the Perspective of Generalization}
\author{
    Xinli Yue\textsuperscript{\rm 1}\equalcontrib\thanks{Work done during Xinli Yue’s internship at WeChat.},
    Jianhui Sun\textsuperscript{\rm 2}\equalcontrib,
    Liangchao Yao\textsuperscript{\rm 2},
    Fan Xia\textsuperscript{\rm 2},
    Yuetang Deng\textsuperscript{\rm 2},
    Tianyi Wang\textsuperscript{\rm 2},
    Lei Li\textsuperscript{\rm 2},
    Fengyun Rao\textsuperscript{\rm 2},
    Jing Lv\textsuperscript{\rm 2},
    Qian Wang\textsuperscript{\rm 1},
    Lingchen Zhao\textsuperscript{\rm 1}\thanks{Corresponding author. lczhaocs@whu.edu.cn},
}
\affiliations{
    \textsuperscript{\rm 1}School of Cyber Science and Engineering, Wuhan University, Wuhan, China\\
    \textsuperscript{\rm 2}WeChat, Tencent Inc, Guangzhou, China\\
}

\usepackage{bibentry}

\begin{document}

\maketitle

\begin{abstract}
The increasing popularity of short video platforms such as YouTube Shorts, TikTok, and Kwai has led to a surge in User-Generated Content (UGC), which presents significant challenges for the generalization performance of Video Quality Assessment (VQA) tasks. These challenges not only affect performance on test sets but also impact the ability to generalize across different datasets. While prior research has primarily focused on enhancing feature extractors, sampling methods, and network branches, it has largely overlooked the generalization capabilities of VQA tasks. In this work, we reevaluate the VQA task from a generalization standpoint. We begin by analyzing the weight loss landscape of VQA models, identifying a strong correlation between this landscape and the generalization gaps. We then investigate various techniques to regularize the weight loss landscape. Our results reveal that adversarial weight perturbations can effectively smooth this landscape, significantly improving the generalization performance, with cross-dataset generalization and fine-tuning performance enhanced by up to 1.8\% and 3\%, respectively. Through extensive experiments across various VQA methods and datasets, we validate the effectiveness of our approach. Furthermore, by leveraging our insights, we achieve state-of-the-art performance in Image Quality Assessment (IQA) tasks. Our code is available at: https://github.com/XinliYue/VQA-Generalization.
\end{abstract}

%

\section{Introduction}

With the rapid rise of short video platforms, User-Generated Content (UGC) has experienced explosive growth. This content not only increases significantly in quantity but also becomes increasingly diverse in form and substance. Concurrently, advancements in video capture devices have led to continuous improvements in video resolution. These developments present unprecedented challenges for Video Quality Assessment (VQA) tasks, particularly in terms of model generalization performance. These challenges are evident not only in the generalization performance on test sets but also in zero-shot cross-dataset generalization and fine-tuned generalization on target datasets.

Although substantial research has been dedicated to enhancing the performance of VQA models, most efforts have primarily focused on improving feature extractors. For instance, the evolution has progressed from early handcrafted features to Convolutional Neural Networks (CNNs)~\cite{vsfa,simplevqa} and more recently to Vision Transformers (ViTs)~\cite{fastvqa,fastervqa,dover}, and even large multi-modality models (LMMs)~\cite{qlign}. This includes the transition from using image-pretrained models to video-pretrained models. Additionally, some studies have attempted to improve model performance through advanced sampling methods, such as grid-based segment sampling~\cite{fastvqa, fastervqa} and segment pyramid sampling~\cite{sama}, or by increasing the number of model branches~\cite{rich, dover, zoomvqa}. However, research that considers VQA tasks from a generalization perspective remains relatively scarce. Therefore, our research aims to systematically investigate VQA tasks from a generalization standpoint, with the goal of enhancing the generalization performance of VQA models, including zero-shot generalization, to address the significant challenges currently faced by VQA tasks.

In classic classification tasks, extensive research~\cite{largebatch, exploring, visual, diametrical, awp, sam} has already delved into generalization, demonstrating both theoretically and empirically that the smoothness of the weight loss landscape (which is the loss change with respect to the weight), particularly the flatness of minima, is closely associated with generalization. Thus, we take the weight loss landscape as our starting point, first visualizing the weight loss landscapes of various VQA models to study their relationship with generalization performance. We find that the smoothness of the weight loss landscape and the generalization gap (i.e., the performance difference between training and test sets) are closely related across different VQA models, including models trained with the same method at different epochs and models trained with different methods. Consequently, to enhance the generalization performance of VQA models, we focus on the smoothness of the weight loss landscape, attempting to implicitly or explicitly regularize the smoothness of the weight loss landscape while minimizing both the loss value and the sharpness of the loss.

We observed that $L_2$ regularization has a minimal effect on regularizing the weight loss landscape. While data augmentation can smooth the weight loss landscape and reduce the generalization gap, it does not improve generalization performance on the test set. This finding contradicts previous claims that the smoothness of the weight loss landscape is correlated with generalization performance. We further experimented with weight perturbation, specifically adding slight perturbations to model weights during training to reduce the sensitivity of the loss to these perturbations. Our results indicate that random weight perturbations, although capable of smoothing the weight loss landscape, do not enhance generalization performance. In contrast, adversarial weight perturbations not only smooth the loss landscape but also significantly improve generalization performance.

Our contributions can be summarized as follows:
\begin{itemize}
    \item We have discovered a close relationship between the training and testing generalization gap of current VQA methods and the weight loss landscape. Specifically, the smoother the loss landscape, the smaller the generalization gap.
    \item We explored various regularization methods to simultaneously optimize loss minima and loss smoothness during training, finding that adversarial weight perturbations yield the best results.
    \item We conducted extensive experiments across different VQA methods and datasets, providing empirical evidence that adversarial weight perturbations significantly enhance the generalization performance of VQA models, with performance enhanced by up to 3\%.
\end{itemize}

\section{Related Work}
\subsection{Video Quality Assessment (VQA)}
TLVQM~\cite{tlvqm} combines spatially high-complexity and temporally low-complexity handcrafted features, while VIDEVAL~\cite{videval} integrates various handcrafted features to model diverse real-world distortions. However, the factors affecting video quality are highly complex, and these handcrafted features fail to capture them comprehensively. Consequently, more recent efforts have employed deep neural networks to address this issue. VSFA~\cite{vsfa} uses features extracted from pre-trained models. SimpleVQA~\cite{simplevqa} extracts spatial features for spatial distortions and spatiotemporal features for motion distortions to evaluate video quality. FAST-VQA~\cite{fastvqa, fastervqa} proposes grid-based patch sampling, preserving local quality, unbiased global quality, and contextual relationships. Building on this, SAMA~\cite{sama} introduces a patch pyramid to retain both local and global views and proposes a masking strategy to reduce the pyramid to the same input size as FAST-VQA. Q-ALIGN~\cite{qlign} leverages Large Multimodal Models (LMMs) to simulate the scoring process of human raters, utilizing text-defined rating levels to instruct LMMs in the evaluation process. However, these works primarily consider the VQA task from the perspective of feature extraction or data preprocessing.

\subsection{Generalization}

Compared to image-related tasks such as image classification and Image Quality Assessment (IQA), training for VQA is more challenging. On one hand, videos have an additional temporal dimension compared to images, necessitating more complex networks to extract useful features. On the other hand, due to the high cost of annotation, VQA datasets are relatively scarce~\cite{ntire}. The increasing volume of user-generated content (UGC) videos presents unprecedented challenges for VQA tasks. Therefore, training VQA models with stronger generalization capabilities and robust zero-shot evaluation across datasets remains an unresolved issue. 
Regarding generalization performance, both natural~\cite{largebatch, exploring, visual, sam} and adversarial settings~\cite{awp} in classification tasks have been extensively explored. Many studies investigate generalization performance from the perspective of the weight loss landscape, which examines the variation of loss with respect to changes in weights, revealing the geometric characteristics of the loss landscape around the model weights.

\subsection{Weight Perturbation}

The geometric shape of the weight loss landscape, particularly the flatness of its minima, has been extensively studied in relation to generalization~\cite{exploring, visual}. Consequently, several works~\cite{awp, sam, revisiting} have focused on effectively regularizing the weight loss landscape to enhance generalization performance. One approach defines the optimization objective as a min-max problem, aiming to minimize the training loss under adversarial weight perturbations (AWP). This method, also known as Sharpness-Aware Minimization (SAM)~\cite{sam}, improves model generalization by simultaneously minimizing both the loss value and its sharpness. Another approach, exemplified by LPF-SGD~\cite{lpfsgd}, seeks to minimize the expected training loss through random weight perturbations (RWP).

\section{Relationship Between Weight Loss Landscape and VQA Generalization Gap}

Previous studies~\cite{exploring, visual, awp, sam, diametrical} have demonstrated a close relationship between the weight loss landscape and the generalization gap in classification tasks under both natural and adversarial settings. Specifically, a smoother weight loss landscape, characterized by a flatter loss surface near the minima, correlates with a smaller generalization gap. However, in the context of VQA tasks, research on the relationship between the weight loss landscape and the VQA generalization gap remains limited. Therefore, in this section, we first propose a method for visualizing the weight loss landscape in VQA tasks and then investigate the relationship between the weight loss landscape and the VQA generalization gap.

\subsection{Weight Loss Landscape Visualization}

We opted to use filter normalization~\cite{visual} for visualizing the weight loss landscape. Taking the FAST-VQA~\cite{fastvqa} as an example, we visualize the weight loss landscape by plotting the loss changes as the weights $\mathbf{w}$ move along a random direction 
$\mathbf{d}$, where the magnitude of $\mathbf{d}$ is $\alpha$:

\begin{equation}
g(\alpha)=\frac{1}{n} \sum_{i=1}^n \ell\left(\mathbf{f}_{\mathbf{w}+\alpha \mathbf{d}}\left(\mathbf{x}_i\right), y_i\right)
\end{equation}
\begin{equation}
\ell=\ell_{\mathrm{PLCC}}+\beta \cdot \ell_{\mathrm{RANK}}
\label{eq_loss}
\end{equation}
\begin{equation}
\ell_{\mathrm{RANK}}=\sum_{i, j} \max \left(\left(\mathbf{f}_{\mathbf{w}}\left(\mathbf{x}_i\right)-\mathbf{f}_{\mathbf{w}}\left(\mathbf{x}_j\right)\right) \operatorname{sgn}\left(s_j-s_i\right), 0\right)
\end{equation}
\begin{equation}
\ell_{\mathrm{PLCC}}=\left(1-\frac{\left(\mathbf{f}_{\mathbf{w}}(\mathbf{x})-\overline{\mathbf{f}_{\mathbf{w}}(\mathbf{x})}\right) \cdot(s-\bar{s})}{\left\|\mathbf{f}_{\mathbf{w}}(\mathbf{x})-\overline{\mathbf{f}_{\mathbf{w}}(\mathbf{x})}\right\|_2\|s-\bar{s}\|_2}\right) / 2  
\end{equation}
where $\mathbf{d}$ is sampled from a Gaussian distribution and normalized using filter normalization as $\mathbf{d}_{i, j} \leftarrow \frac{\mathbf{d}_{i, j}}{\left\|\mathbf{d}_{i, j}\right\|_F}\left\| \mathbf{w}_{i, j} \right\|_F$ (where $\mathbf{d}_{i, j}$ represents the $j$-th filter of the $i$-th layer of $\mathbf{d}$, and $\left\|\cdot\right\|_F$ denotes the Frobenius norm), to eliminate the scale invariance of DNNs. The loss $\ell$ employed by FAST-VQA consists of the monotonicity loss $\ell_{\text {RANK }}$ and the linear fusion loss $\ell_{\text {PLCC }}$. For simplicity, we omit the parameters $\mathbf{f_w}, \textbf{x}_i, s_i$ in the subsequent loss function descriptions. Here, $\mathbf{f_w}$ denotes the network parameterized by $\mathbf{w}$, $\mathbf{x}_i$ represents the video sample, and $s_i$ is the ground-truth quality score of the video. The hyperparameter $\beta$ measures the weight of $\ell_{\text {RANK }}$, and sgn is the sign function.

\subsection{Generalization Gap During Training}

\begin{figure}[ht]  
\centering
\subfloat[]
{\includegraphics[width=.48\linewidth]{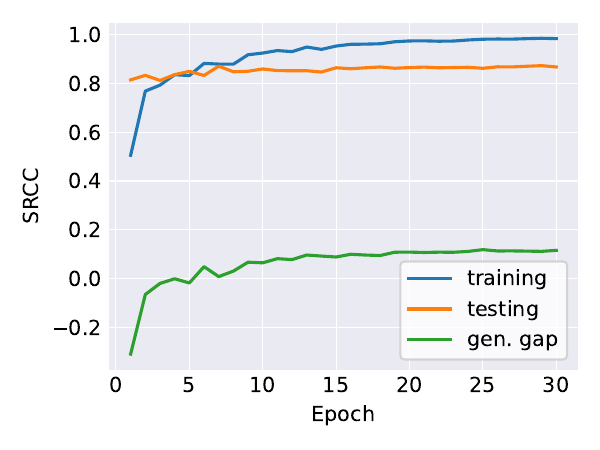}}
\subfloat[]
{\includegraphics[width=.48\linewidth]{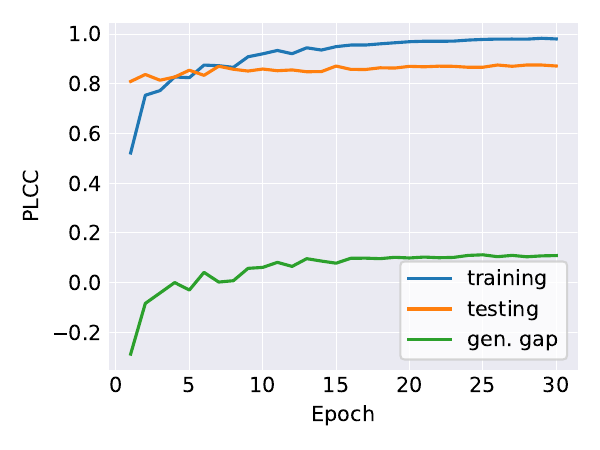}} \\
\subfloat[]
{\includegraphics[width=.48\linewidth]{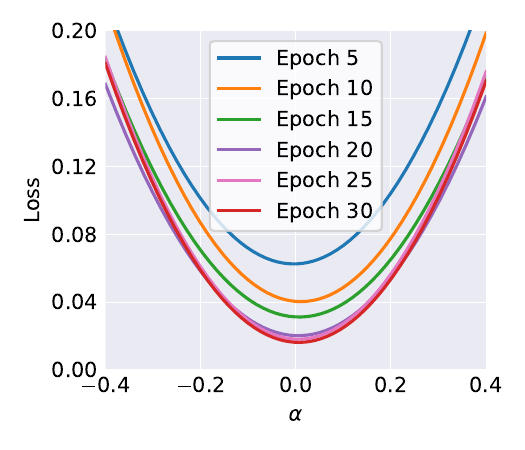}}
\subfloat[]
{\includegraphics[width=.48\linewidth]{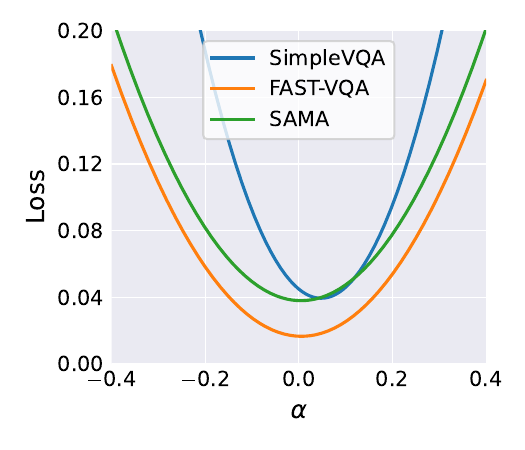}} \\
\caption{The training results of various methods on the KoNViD-1k dataset~\cite{konvid}. (a) and (b) illustrate the performance metrics of FAST-VQA~\cite{fastvqa} on both the training and testing sets across different epochs, as well as the generalization gap. (c) compares the weight loss landscape of FAST-VQA at different epochs. (d) contrasts the weight loss landscapes of different VQA methods.}
\label{fig_epoch}
\end{figure}

Using the KoNViD-1k~\cite{konvid} dataset as an example, we plotted the SRCC and PLCC on both the training and test sets, as well as the SRCC and PLCC generalization gap (refer to the experimental setup for SRCC and PLCC) for FAST-VQA~\cite{fastvqa} during the training process, as shown in Figures \ref{fig_epoch}(a) and \ref{fig_epoch}(b). Additionally, we visualized the weight loss landscape of models at different epochs in Figure \ref{fig_epoch}(c). It can be observed that as training progresses, the generalization gap consistently increases, and the weight loss landscape becomes increasingly sharp. This evidence supports the close relationship between the smoothness of the weight loss landscape and the generalization gap, indicating that a smaller generalization gap corresponds to a smoother weight loss landscape.

\subsection{Generalization Gap Across Different Methods}

After exploring the generalization gap during the training process, we further investigated whether the relationship between the weight loss landscape and the generalization gap persists across different VQA methods. We examined three classical single-branch VQA methods: SimpleVQA~\cite{simplevqa}, FAST-VQA~\cite{fastvqa}, and SAMA~\cite{sama}. SimpleVQA employs L1 loss and RANK loss as its loss functions; thus, we replaced $\ell_{\text {PLCC }}$ with $\ell_{\mathrm{L} 1}$ in the formula while keeping other components consistent to visualize the weight loss landscape of SimpleVQA. Table \ref{tab_method} presents the generalization gaps of different methods, and Figure \ref{fig_epoch}(d) illustrates their respective weight loss landscapes. It can be observed that SimpleVQA exhibits the largest generalization gap and the sharpest weight loss landscape. In comparison, SAMA has a slightly smoother loss landscape than FAST-VQA and a correspondingly smaller generalization gap.

From this, it can be observed that, similar to the conclusions drawn by works such as ~\cite{awp}, there is a close relationship between the smoothness of the weight loss landscape and the generalization gap. Intuitively, if we aim to reduce the generalization gap, we can attempt to enhance the smoothness of the weight loss landscape, either implicitly or explicitly.

\begin{table}[t]
\setlength{\tabcolsep}{1.5mm}{
\begin{tabular}{@{}ccccccc@{}}
\toprule[1.0pt]
\multirow{2}{*}{Method} & \multicolumn{3}{c}{SRCC}                         & \multicolumn{3}{c}{PLCC}                         \\ \cmidrule(l){2-4} \cmidrule(l){5-7} 
                        & Train          & Test           & Gap           & Train          & Test           & Gap           \\ \midrule
SimpleVQA               & 0.960          & 0.806          & 0.154          & 0.959          & 0.832          & 0.127          \\
FAST-VQA                & \textbf{0.983} & 0.867          & 0.116          & \textbf{0.980} & 0.872          & 0.109          \\
SAMA                    & 0.978          & \textbf{0.873} & \textbf{0.105} & 0.974          & \textbf{0.873} & \textbf{0.101} \\ \bottomrule[1.0pt]
\end{tabular}
}
\caption{Performance metrics of different VQA methods on the KoNViD-1k~\cite{konvid} training and testing sets, along with the generalization gap.}
\label{tab_method}
\end{table}

\section{Reducing the Generalization Gap}
\subsection{$L_2$ Regularization}
\begin{figure}[ht]
    \centering

    \subfloat[]
    {\includegraphics[width=.48\linewidth]{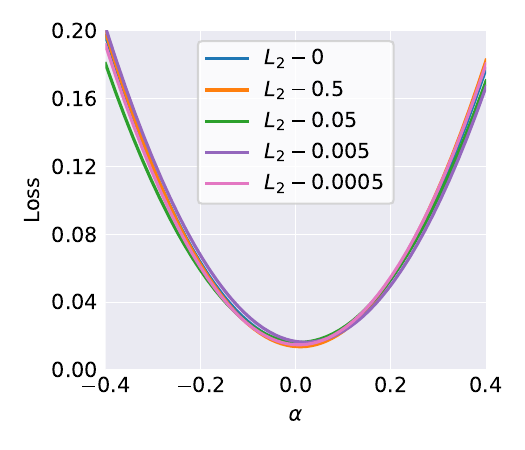}}
    \subfloat[]
    {\includegraphics[width=.48\linewidth]{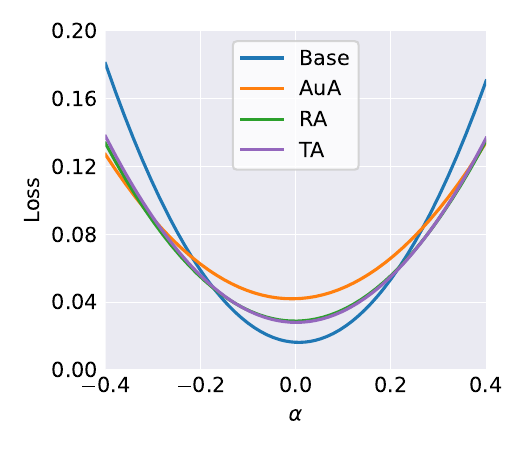}} \\
    \subfloat[]
    {\includegraphics[width=.48\linewidth]{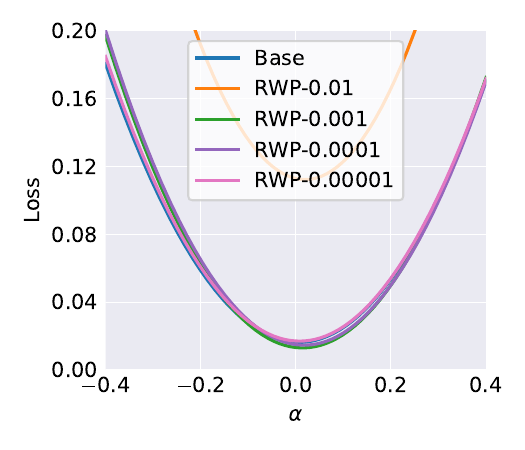}}
    \subfloat[]
    {\includegraphics[width=.48\linewidth]{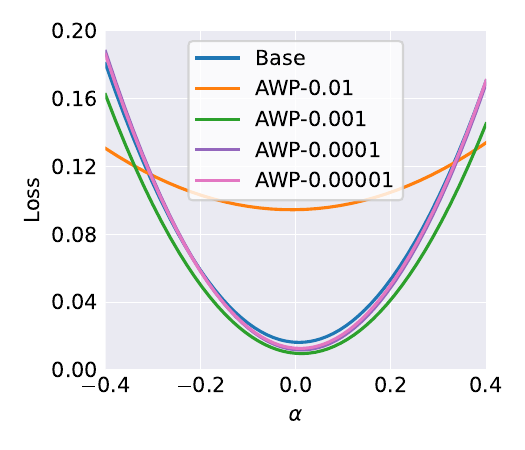}}
    \caption{Weight loss landscape of FAST-VQA on the KoNViD-1k dataset under various modifications. (a) Different $L_2$ regularization parameters $\lambda$, (b) Different data augmentation methods, (c) Different RWP perturbation magnitudes, (d) Different AWP perturbation magnitudes.}
    \label{fig_loss_landscape}

\vspace{-2pt}
\end{figure}

$L_2$ regularization penalizes large weight values, encouraging the model to learn smaller weights. This penalty makes the loss function smoother in the weight space, as larger weights typically lead to abrupt changes in the loss function. Specifically, $L_2$ regularization is achieved by adding a penalty term proportional to the sum of the squared weights to the loss function. Formally, if the original loss function is $\ell$, the loss function with $L_2$ regularization becomes:
\begin{equation}
    \ell_{\text{new}}=\ell+\lambda \sum_i \mathbf{w}_i^2
\end{equation}
By adjusting the parameters of $L_2$ regularization, we can control the intensity of $L_2$ weight regularization. For instance, FAST-VQA uses a default $L_2$ regularization parameter of $\lambda=5e-2$. We adjusted $\lambda$ to values of $\{0, 5e-1,5e-3,5e-4\}$, and observed that the generalization performance of FAST-VQA did not significantly improve compared to the default setting. The weight loss landscape in Figure \ref{fig_loss_landscape}(a) also showed that the smoothness of the loss landscape under different $\lambda$ parameters was quite similar, almost overlapping. Therefore, $L_2$ regularization has a limited effect on increasing the smoothness of the loss landscape and reducing the generalization gap.

\begin{table}[t]
\centering
\setlength{\tabcolsep}{2.2mm}{
\begin{tabular}{@{}ccccccc@{}}
\toprule[1.0pt]
\multirow{2}{*}{Reg.} & \multicolumn{3}{c}{SRCC}                         & \multicolumn{3}{c}{PLCC}                         \\ \cmidrule(l){2-4} \cmidrule(l){5-7}
                      & Train          & Test           & Gap           & Train          & Test           & Gap           \\ \midrule
0                     & 0.983          & 0.871          & 0.112          & 0.982          & 0.874          & 0.108          \\
0.5                   & 0.983          & 0.870          & 0.113          & 0.982          & \textbf{0.876} & \textbf{0.106} \\
0.05                  & \textbf{0.984} & 0.872          & 0.112          & \textbf{0.983} & \textbf{0.876} & 0.108          \\
0.005                 & 0.982          & 0.872          & 0.111          & 0.981          & 0.874          & 0.107          \\
0.0005                & 0.982          & \textbf{0.875} & \textbf{0.108} & 0.980          & 0.875          & \textbf{0.106} \\ \bottomrule[1.0pt]
\end{tabular}
}
\caption{Performance metrics of FAST-VQA~\cite{fastvqa} on the KoNViD-1k training and testing sets with different $L_2$ regularization parameters $\lambda$, along with the generalization gap.}
\label{tab_$L_2$}
\end{table}

\subsection{Data Augmentation}

Data augmentation, a common regularization technique, enhances the diversity of samples to expand the training distribution. When faced with augmented samples, the model must adapt to a broader range of input variations. During training, each sample point encountered by the model is no longer highly concentrated or repetitive but rather more dispersed and covers a wider input space. Consequently, the model is less likely to form sharp responses (i.e., overfit) to specific, frequently occurring sample points, as these points become less frequent or appear in different forms after augmentation. Instead, the model needs to optimize and adapt to a broader set of samples, leading to more uniform behavior across the input space. This results in smoother gradient changes in the loss function, reducing sharp local minima.

We experimented with three different data augmentation strategies, including AutoAugment (AuA)~\cite{aua}, RandAugment (RA)~\cite{ra}, and TrivialAugment (TA)~\cite{ta}. FAST-VQA employs a sampling strategy known as Grid Mini-patch Sampling (GMS), which divides the video into spatially uniform, non-overlapping grids, randomly sampling a mini-patch from each grid, and then stitching these mini-patches together. In our experiments, we found that applying data augmentation after the patching process disrupts the effectiveness of the patches, leading to lower model performance. Therefore, we adopted a strategy of applying data augmentation to the frames before patching. The generalization gap and weight loss landscape are presented in Table \ref{tab_aug} and Figure \ref{fig_loss_landscape}(b), respectively. It is evident that the loss smoothness for all four data augmentation methods is higher than that of the vanilla approach, and the generalization gap is also lower than the baseline. This once again demonstrates the relationship between the weight loss landscape and the generalization gap.

However, despite the lower generalization gap for the three data augmentation methods compared to the vanilla approach, the generalization performance on the test set did not improve. In fact, the generalization performance for RA and TA was even lower than that of the vanilla approach. This indicates that while a smoother weight loss landscape can lead to a smaller generalization gap, it does not necessarily guarantee better generalization performance. 

\begin{table}[t]
\centering
\setlength{\tabcolsep}{2.3mm}{
\begin{tabular}{@{}ccccccc@{}}
\toprule[1.0pt]
\multirow{2}{*}{Aug.} & \multicolumn{3}{c}{SRCC}                         & \multicolumn{3}{c}{PLCC}                         \\ \cmidrule(l){2-4} \cmidrule(l){5-7} 
                      & Train          & Test           & Gap           & Train          & Test           & Gap           \\ \midrule
base                  & \textbf{0.984} & 0.872          & 0.112          & \textbf{0.983} & \textbf{0.876} & 0.108          \\
AuA                   & 0.912          & \textbf{0.875} & \textbf{0.036} & 0.909          & 0.868          & \textbf{0.041} \\
RA                    & 0.953          & 0.860          & 0.093          & 0.950          & 0.861          & 0.089          \\
TA                    & 0.948          & 0.864          & 0.083          & 0.944          & 0.868          & 0.076          \\ \bottomrule[1.0pt]
\end{tabular}
}
\caption{Performance metrics of FAST-VQA with different data augmentation methods on the KoNViD-1k training and testing sets, along with the generalization gap.}
\label{tab_aug}
\end{table}

\subsection{Weight Perturbation}

In addition to implicit methods such as $L_2$ regularization and data augmentation for smoothing the weight loss landscape, weight perturbation is an explicit method for achieving this goal. Weight perturbation involves adding slight perturbations to the weights during training, allowing the loss to decrease in the presence of these perturbations and reducing the sensitivity of the loss to minor changes. This enhances the smoothness of the loss landscape, effectively reducing its sharpness. The sensitivity of the loss to weight perturbations can be formalized as:

\begin{equation}
   \Delta \ell(\mathbf{w}, \boldsymbol{\epsilon})=\ell(\mathbf{w}+\boldsymbol{\epsilon})-\ell(\mathbf{w}) 
\end{equation}

Thus, the optimization objectivemin $\min_{\mathbf{w}}\{\ell(\mathbf{w})\}$ transforms into:
\begin{equation}
    \min_{\mathbf{w}}\{\ell(\mathbf{w})+\Delta \ell(\mathbf{w}, \boldsymbol{\epsilon})\}=\min_{\mathbf{w}}\{\ell(\mathbf{w})+(\ell(\mathbf{w}+\boldsymbol{\epsilon})-\ell\left(\mathbf{w}\right))\}
\end{equation}

\begin{equation}
    \min_{\mathbf{w}}\{\ell(\mathbf{w})+(\ell(\mathbf{w}+\boldsymbol{\epsilon})-\ell(\mathbf{w}))\} \rightarrow \min_{\mathbf{w}} \ell(\mathbf{w}+\boldsymbol{\epsilon})
\end{equation}
where $\boldsymbol{\epsilon}$ represents the perturbation added to the weights. Currently, there are two mainstream methods for generating perturbations $\boldsymbol{\epsilon}$: Random Weight Perturbation (RWP)~\cite{lpfsgd} and Adversarial Weight Perturbation (AWP)~\cite{awp}. To maintain consistency between the two, we only alter the direction of the perturbations while keeping the magnitude constant.

\subsubsection{Perturbation Direction}
The direction of RWP is randomly sampled from a Gaussian distribution, formalized as:
\begin{equation}
    \boldsymbol{\epsilon} \sim \mathcal{N}(\mathbf{0}, I)
\end{equation}

In contrast, the direction of AWP is the one that maximizes the loss, i.e.,
\begin{equation}
    \boldsymbol{\epsilon}={\arg \max } 
 \,\ell(\mathbf{w}+\boldsymbol{\epsilon})
\end{equation}

where $\mathbf{w}$ represents the model weights. AWP injects the worst-case weight perturbation within a small neighborhood around $\mathbf{f}_{\mathbf{w}}$.

\begin{table}[t]
\centering
\setlength{\tabcolsep}{2.1mm}{
\begin{tabular}{@{}ccccccc@{}}
\toprule[1.0pt]
\multirow{2}{*}{RWP} & \multicolumn{3}{c}{SRCC}                         & \multicolumn{3}{c}{PLCC}                         \\ \cmidrule(l){2-4} \cmidrule(l){5-7} 
                     & Train          & Test           & Gap           & Train          & Test           & Gap           \\ \midrule
0                    & \textbf{0.984} & 0.872          & 0.112          & \textbf{0.983} & \textbf{0.876} & 0.108          \\
0.01                 & 0.886          & 0.839          & \textbf{0.047} & 0.883          & 0.840          & \textbf{0.043} \\
0.001                & 0.975          & \textbf{0.875} & 0.099          & 0.974          & 0.874          & 0.100          \\
0.0001               & 0.976          & 0.870          & 0.106          & 0.975          & 0.870          & 0.105          \\
0.00001              & 0.977          & 0.869          & 0.108          & 0.975          & 0.868          & 0.107          \\ \bottomrule[1.0pt]
\end{tabular}
}
\caption{Performance metrics of FAST-VQA with RWP improvements of varying perturbation magnitudes on the KoNViD-1k training and testing sets, along with the generalization gap.}
\label{tab_rwp}
\end{table}

\begin{table}[t]
\centering
\setlength{\tabcolsep}{1.9mm}{
\begin{tabular}{@{}ccccccc@{}}
\toprule[1.0pt]
\multirow{2}{*}{AWP} & \multicolumn{3}{c}{SRCC}                          & \multicolumn{3}{c}{PLCC}                          \\ \cmidrule(l){2-4} \cmidrule(l){5-7} 
                     & Train          & Test           & Gap            & Train          & Test           & Gap            \\ \midrule
0                    & \textbf{0.984} & 0.872          & 0.112           & \textbf{0.983} & 0.876          & 0.108           \\
0.01                 & 0.784          & 0.800          & \textbf{-0.016} & 0.780          & 0.817          & \textbf{-0.037} \\
0.001                & 0.972          & 0.882          & 0.090           & 0.969          & 0.884          & 0.085           \\
0.0001               & 0.975          & \textbf{0.884} & 0.091           & 0.973          & \textbf{0.890} & 0.083           \\
0.00001              & 0.982          & 0.877          & 0.105           & 0.978          & 0.884          & 0.095           \\ \bottomrule[1.0pt]
\end{tabular}
}
\caption{Performance metrics of FAST-VQA with AWP improvements of varying perturbation magnitudes on the KoNViD-1k training and testing sets, along with the generalization gap.}
\label{tab_awp}
\vspace{-8pt}
\end{table}

\subsubsection{Perturbation Magnitude}
Based on the direction of the weight perturbation, it is essential to determine the magnitude of the perturbation to be injected. Following AWP, we constrain the weight perturbation $\boldsymbol{\epsilon}_l$ relative to the magnitude of the weights in the $l$-th layer $\mathbf{w}_l$:
\begin{equation}
    \left\|\boldsymbol{\epsilon}_l\right\| \leq \gamma\left\|\mathbf{w}_l\right\|
\end{equation}
where $\gamma$ is the constraint on the magnitude of the weight perturbation.

\begin{algorithm}[t]
\caption{FAST-VQA-WP}
\label{alg:algorithm}
\textbf{Input}: model $\mathbf{f_w}$,batch size $m$,learning rate $\eta_1$, perturbation constraint $\gamma$, perturbation step size $\eta_2$, perturbation steps $K$ \\
\textbf{Output}: model $\mathbf{f_w}$
\begin{algorithmic}[1] 
\REPEAT
\STATE Read mini-batch $B={\mathbf{x}_1,\ldots,\mathbf{x}_m }$ from training set
\IF{RWP}
\STATE $\boldsymbol{\epsilon} \sim \mathcal{N}(\mathbf{0}, I)$
\ELSE 
\FOR{$k=1,\ldots,K$}
\STATE $\boldsymbol{\epsilon} \leftarrow \Pi_\gamma\left(\boldsymbol{\epsilon}+\eta_2 \frac{\nabla_{\boldsymbol{\epsilon}} \frac{1}{m} \sum_{i=1}^m \ell\left(\mathbf{f}_{\mathbf{w}+\boldsymbol{\epsilon}}\left(\mathbf{x}_i\right), s_i\right)}{\left\|\nabla_{\boldsymbol{\epsilon}} \frac{1}{m} \sum_{i=1}^m \ell\left(\mathbf{f}_{\mathbf{w}+\boldsymbol{\epsilon}}\left(\mathbf{x}_i\right), s_i\right)\right\|}\|\mathbf{w}\|\right)$
\ENDFOR
\ENDIF \\
\STATE
$\mathbf{w} \leftarrow(\mathbf{w}+\boldsymbol{\epsilon})-\eta_2 \nabla_{\mathbf{w}+\boldsymbol{\epsilon}} \frac{1}{m} \sum_{i=1}^m \ell\left(\mathbf{f}_{\mathbf{w}+\boldsymbol{\epsilon}}\left(\mathbf{x}_i, s_i\right)\right)-\boldsymbol{\epsilon}$
\UNTIL{training converged}
\end{algorithmic}
\end{algorithm}
\vspace{-8pt}

\subsubsection{Perturbation Calculation}
Once the direction and magnitude of the perturbation are determined, the next step is to compute the perturbation. Since RWP involves random sampling, its computation is relatively straightforward. In contrast, AWP requires calculating the gradient of the weights $\mathbf{w}$ with respect to the loss function $\ell$, and then adding perturbations in the direction of the gradient ascent.
\begin{equation}
    \boldsymbol{\epsilon} \leftarrow \Pi_\gamma\left(\boldsymbol{\epsilon}+\eta_2 \frac{\nabla_{\boldsymbol{\epsilon}} \frac{1}{m} \sum_{i=1}^m \ell\left(\mathbf{f}_{\mathbf{w}+\boldsymbol{\epsilon}}\left(\mathbf{x}_i\right), s_i\right)}{\left\|\nabla_{\boldsymbol{\epsilon}} \frac{1}{m} \sum_{i=1}^m \ell\left(\mathbf{f}_{\mathbf{w}+\boldsymbol{\epsilon}}\left(\mathbf{x}_i\right), s_i\right)\right\|}\|\mathbf{w}\|\right)
\end{equation}

where $m$ is the batch size, $\eta_2$ is the learning rate, and $\boldsymbol{\epsilon}$ is updated layer by layer. The perturbation $\boldsymbol{\epsilon}$ can be solved using either a single-step or multi-step method.

\begin{table*}[t]
\centering
\setlength{\tabcolsep}{2.0mm}{
\begin{tabular}{@{}cccccccccccc@{}}
\toprule[1.0pt]
\multicolumn{2}{c}{Trained on $\text{LSVQ}_\text{train}$}              & \multicolumn{2}{c}{$\text{LSVQ}_\text{test}$}  & \multicolumn{2}{c}{$\text{LSVQ}_\text{1080p}$} & \multicolumn{2}{c}{KoNViD-1k}   & \multicolumn{2}{c}{LIVE-VQC}    \\ \cmidrule(l){1-2} \cmidrule(l){3-4} \cmidrule(l){5-6} \cmidrule(l){7-8} \cmidrule(l){9-10} 
Type                           & Method          & SRCC           & PLCC           & SRCC           & PLCC           & SRCC           & PLCC           & SRCC           & PLCC           \\ \midrule
\multirow{2}{*}{hand-crafted}  & TLVQM~\cite{tlvqm}           & 0.772          & 0.774          & 0.589         & 0.616          & 0.732          & 0.724          & 0.670          & 0.691          \\
                               & VIDEVAL~\cite{videval}         & 0.795          & 0.783          & 0.545          & 0.554          & 0.751          & 0.741          & 0.630          & 0.640          \\ \midrule
\multirow{6}{*}{multi-branch}  & PVQ (w/o patch)~\cite{pvq} & 0.814          & 0.816          & 0.686          & 0.708          & 0.781          & 0.781          & 0.747          & 0.776          \\
                               & PVQ (w patch)~\cite{pvq}   & 0.827          & 0.828          & 0.711          & 0.739          & 0.791          & 0.795          & 0.770          & 0.807          \\
                               & BVQA-2022~\cite{bvqa}       & 0.852          & 0.855          & 0.771          & 0.782          & 0.834          & 0.837          & 0.816          & 0.824          \\
                               & DSD-PRO~\cite{dsdpro}         & 0.875          & 0.875          & 0.765          & 0.793          & 0.865          & 0.856          & \textbf{0.838} & 0.849          \\
                               & ZoomVQA~\cite{zoomvqa}         & 0.886          & 0.879          & \textbf{0.799} & 0.819          & 0.877          & 0.875          & 0.814          & 0.833          \\
                               & DOVER~\cite{dover}           & \textbf{0.888} & \textbf{0.889} & 0.795          & \textbf{0.830} & \textbf{0.884} & \textbf{0.883} & 0.832          & \textbf{0.855} \\ \midrule
LMMs                           & Q-ALIGN~\cite{qlign}         & 0.883          & 0.882          & 0.797          & \textbf{0.830}          & 0.865          & 0.877          & 0.778              & 0.821              \\ \midrule
\multirow{6}{*}{single-branch} & SimpleVQA~\cite{simplevqa}       & 0.856          & 0.853          & 0.740          & 0.783          & 0.855          & 0.856          & 0.750          & 0.805          \\
                               & +AWP & \textbf{0.864} & \textbf{0.863} & \textbf{0.747} & \textbf{0.792} & \textbf{0.863} & \textbf{0.863} & \textbf{0.768} & \textbf{0.815} \\ \cmidrule(l){2-10}
                               & FAST-VQA~\cite{fastvqa}        & 0.876          & 0.877          & \textbf{0.779} & \textbf{0.814} & 0.859          & 0.855          & 0.823          & 0.844          \\
                               & +AWP & \textbf{0.879} & \textbf{0.881} & \textbf{0.779}          & 0.813          & \textbf{0.871} & \textbf{0.866} & \textbf{0.839} & \textbf{0.851} \\ \cmidrule(l){2-10}
                               & SAMA~\cite{sama}            & 0.879          & 0.880          & 0.782          & 0.820          & 0.873          & 0.871          & 0.828          & 0.845          \\
                               & +AWP & \textbf{0.883} & \textbf{0.883} & \textbf{0.787} & \textbf{0.824} & \textbf{0.877} & \textbf{0.876} & \textbf{0.834} & \textbf{0.851} \\  \bottomrule[1.0pt]
\end{tabular}
}
\caption{Comparison of the generalization performance of models trained on LSVQ using different methods across various datasets.}
\label{tab_main}
\vspace{-8pt}
\end{table*}

\begin{table*}[t]
\centering
{
\begin{tabular}{@{}cccccccc@{}}
\toprule[1.0pt]
\multirow{2}{*}{Type}          & \multirow{2}{*}{Method} & \multicolumn{2}{c}{KoNViD-1k}   & \multicolumn{2}{c}{LIVE-VQC}    & \multicolumn{2}{c}{YouTube-UGC} \\ \cmidrule(l){3-4} \cmidrule(l){5-6} \cmidrule(l){7-8} 
                               &                         & SRCC           & PLCC           & SRCC           & PLCC           & SRCC           & PLCC           \\ \cmidrule(r){1-8}
\multirow{2}{*}{hand-crafted}  & TLVQM~\cite{tlvqm}                   & 0.773          & 0.768          & 0.799          & 0.803          & 0.669          & 0.   659       \\
                               & VIDEVAL~\cite{videval}                 & 0.783          & 0.780          & 0.752          & 0.751          & 0.779          & 0.773          \\ \midrule
\multirow{4}{*}{multi-branch}  & PVQ~\cite{pvq}                     & 0.791          & 0.786          & 0.827          & 0.837          & -              & -              \\
                               & BVQA-2022~\cite{bvqa}               & 0.834          & 0.836          & 0.834          & 0.842          & 0.818          & 0.826          \\
                               & DSD-PRO~\cite{dsdpro}                 & 0.861          & 0.861          & 0.875          & 0.862          & 0.841          & 0.836          \\
                               & DOVER~\cite{dover}                   & 0.909          & 0.906          & 0.860          & 0.875          & 0.890          & 0.891          \\ \midrule
\multirow{6}{*}{single-branch} & SimpleVQA~\cite{simplevqa}               & 0.806          & 0.832          & 0.639          & 0.710          & 0.840          & 0.854          \\
                               & +AWP         & \textbf{0.828} & \textbf{0.843} & \textbf{0.641} & \textbf{0.729} & \textbf{0.848} & \textbf{0.858} \\ \cmidrule(l){2-8}
                               & FAST-VQA~\cite{fastvqa}                & 0.887          & 0.896          & 0.852          & 0.876          & 0.855          & 0.852          \\
                               & +AWP         & \textbf{0.893} & \textbf{0.902} & \textbf{0.863} & \textbf{0.886} & \textbf{0.885} & \textbf{0.882} \\ \cmidrule(l){2-8}
                               & SAMA~\cite{sama}                   & 0.892          & 0.892          & 0.855          & 0.883          & 0.881          & 0.877          \\
                               & +AWP         & \textbf{0.911} & \textbf{0.903} & \textbf{0.860} & \textbf{0.887} & \textbf{0.886} & \textbf{0.883} \\ \bottomrule[1.0pt]
\end{tabular}
}
\caption{Comparison of fine-tuning performance of different methods. Due to the suboptimal fine-tuning results of SimpleVQA, the results provided here are from models trained independently.}
\label{tab_finetune}
\end{table*}

\begin{table}[t]
\centering
{
\begin{tabular}{@{}ccccc@{}}
\toprule[1.0pt]
\multirow{2}{*}{Method}             & \multicolumn{2}{c}{KonIQ}       & \multicolumn{2}{c}{SPAQ}        \\ \cmidrule(l){2-3} \cmidrule(l){4-5} 
                                    & SRCC           & PLCC           & SRCC           & PLCC           \\ \midrule
HyperIQA                            & 0.906          & 0.917          & 0.916          & 0.919          \\
MUSIQ                               & 0.916          & 0.928          & 0.917          & 0.921          \\
IEIT                                & 0.892          & 0.916          & 0.917          & 0.921          \\
CVC-IQA                             & 0.915          & 0.941          & 0.925          & 0.929          \\
VT-IQA                              & 0.924          & 0.938          & 0.920          & 0.925          \\
DAC NN                              & 0.901          & 0.912          & 0.915          & 0.921          \\
SARQUE                              & 0.901          & 0.923          & 0.918          & 0.922          \\
CONTRIQUE                           & 0.894          & 0.906          & 0.914          & 0.919          \\
VCRNet                              & 0.894          & 0.909          & -              & -              \\
DEIQT                               & 0.921          & 0.934          & 0.919          & 0.923          \\
Re-IQA                              & 0.923          & 0.914          & 0.918          & 0.925          \\
LIQE                                & 0.919          & 0.908          & -              & -              \\
QPT                                 & 0.927          & 0.941          & 0.925          & 0.928          \\ \midrule
SAMA                                & 0.930          & 0.942          & 0.925          & 0.929          \\
+AWP & \textbf{0.934} & \textbf{0.946} & \textbf{0.927} & \textbf{0.930} \\ \bottomrule[1.0pt]
\end{tabular}
}
\caption{Performance comparison of different methods on the KonIQ and SPAQ datasets.}
\label{tab_iqa}
\vspace{-4pt}
\end{table}

\subsubsection{Model Training}
Algorithm \ref{alg:algorithm} presents the weight perturbation variant of FAST-VQA. For the weight perturbation variants of other VQA methods, such as SimpleVQA and SAMA, please refer to the pseudocode provided in the appendix.

\subsubsection{Results of Weight Perturbation}

The results of RWP's generalization gap are presented in Table \ref{tab_rwp}, and the weight loss landscape is illustrated in Figure \ref{fig_loss_landscape}(c). It can be observed that with a perturbation magnitude of 0.01, RWP has a noticeable smoothing effect on the weight loss landscape and significantly reduces the generalization gap. However, the generalization performance declines considerably, indicating that RWP is not effective in enhancing generalization performance.

The results of AWP are shown in Table \ref{tab_awp} and Figure \ref{fig_loss_landscape}(d). It is evident that with a perturbation magnitude of 0.001, 0.0001, and 0.00001, AWP not only reduces the generalization gap and smooths the weight loss landscape but also improves generalization performance. However, further increasing the perturbation magnitude, while continuing to reduce the generalization gap, leads to a decline in generalization performance. This suggests that excessive perturbation may make the model's learning process more challenging, thereby adversely affecting its generalization capability.

\section{Experiments}
\subsection{Experimental Setup}
\subsubsection{Baseline Methods} We consider handcrafted feature-based methods such as TLVQM~\cite{tlvqm} and VIDEVAL~\cite{videval}; multi-branch methods including PVQ~\cite{pvq}, BVQA-2022~\cite{bvqa}, DSD-PRO~\cite{dsdpro}, ZoomVQA~\cite{zoomvqa}, and DOVER~\cite{dover}; LMMs based method Q-ALIGN~\cite{qlign} and single-branch methods like SimpleVQA~\cite{simplevqa}, FAST-VQA~\cite{fastervqa}, and SAMA~\cite{sama}.

\subsubsection{Training Setup} For the baseline methods, we strictly adhere to the experimental settings of the original methods. For our versions, namely SimpleVQA-AWP, FAST-VQA-AWP, and SAMA-AWP, we also rigorously follow the experimental settings of their vanilla counterparts. For all experiments involving AWP, the perturbation step
$K$ are set to 1, and the perturbation magnitude $\gamma$ is set to 0.0001. During the reproduction of SAMA, we observed significant discrepancies between our results and those reported in the original SAMA paper. Therefore, we rely on our reproduced results for further analysis.

\subsubsection{Training and Benchmark Datasets} Following FAST-VQA~\cite{fastvqa}, we utilize the large-scale $\text{LSVQ}_\text{train}$ dataset~\cite{pvq}, which comprises 28,056 videos, to train VQA models. For evaluation, we select four test sets to assess the models trained on LSVQ. The first two test sets, $\text{LSVQ}_\text{test}$ and $\text{LSVQ}_\text{1080p}$, are official internal test subsets of LSVQ. Additionally, we evaluate the cross-dataset generalization capability of VQA models using three VQA benchmark datasets: KoNViD-1k~\cite{konvid} , LIVE-VQC~\cite{livevqc} and YouTubeUGC~\cite{youtubeugc}.

\subsubsection{Evaluation Metrics} Following SimpleVQA~\cite{simplevqa}, we employ two standard metrics to evaluate the performance of VQA models: the Pearson Linear Correlation Coefficient (PLCC) and the Spearman Rank-Order Correlation Coefficient (SRCC). PLCC reflects the linearity of the VQA algorithm's predictions, while SRCC indicates the monotonicity of the predictions. An excellent VQA model should achieve SRCC and PLCC values close to 1.

\subsection{Experimental Results}
By comparing the vanilla methods with our proposed weight perturbation versions in Table \ref{tab_main}, we observe significant improvements in both SRCC and PLCC metrics. Notably, the zero-shot cross-dataset generalization capability shows a marked enhancement. For instance, our method improves the SRCC performance on LIVE-VQC by 1.8\% compared to SimpleVQA. 
Moreover, after introducing weight perturbations, FAST-VQA demonstrates comparable performance to its improved variant, SAMA, under certain experimental settings, and in some cases, even surpasses SAMA.

The fine-tuning results are presented in Table \ref{tab_finetune}, it can be observed that the test results of models fine-tuned with weight perturbation outperform the vanilla versions, particularly for FAST-VQA, which achieved up to a 3\% improvement. In some metrics, the performance is even comparable to SAMA. Moreover, the improved SAMA results are on par with the multi-branch method DOVER~\cite{dover}.

\subsection{Experiments on IQA}
To further validate the effectiveness of adversarial weight perturbation in enhancing generalization performance, we conducted experiments on the Image Quality Assessment (IQA) task. Following SAMA~\cite{sama}, we selected 13 IQA algorithms for performance comparison, including HyperIQA~\cite{hyperiqa}, MUSIQ~\cite{musiq}, IEIT~\cite{ieit}, CVC-IQA~\cite{cvciqa}, VT-IQA~\cite{vtiqa}, DACNN~\cite{dacnn}, SARQUE~\cite{sarque}, CONTRIQUE~\cite{contrique}, VCRNet~\cite{vcrnet}, DEIQT~\cite{deiqt}, Re-IQA~\cite{reiqa}, LIQE~\cite{liqe}, and QPT~\cite{qpt}. We chose two IQA datasets, KonIQ~\cite{koniq} and SPAQ~\cite{spaq}, for our experiments. The results indicate that the improved SAMA outperforms the original SAMA and achieves state-of-the-art (SOTA) performance among the various IQA algorithms, further substantiating our findings.

\section{Conclusion}

In this paper, we investigate the VQA task from a generalization perspective. First, we utilize weight loss landscape visualization to demonstrate the close relationship between the weight loss landscape and the generalization gap in VQA. Consequently, we explore various regularization methods to smooth the weight loss landscape. Our findings reveal that adversarial weight perturbation significantly enhances the generalization performance of VQA. We validate our findings through experiments conducted on different datasets and using various VQA methods.

\clearpage
\bibliography{aaai25}

\clearpage
\appendix

\section{Algorithms}
\subsection{SimpleVQA-AWP}
Algorithm \ref{alg_simplevqa} outlines the training process for SimpleVQA-AWP. The training loss $\ell_s$ for SimpleVQA is computed as follows:

\begin{equation}
    \ell_\text{MAE}=\frac{1}{n} \sum_{i=1}^n\left|\mathbf{f}_{\mathbf{w}}\left(\mathbf{x}_i\right)-s_i\right|
\end{equation}

\begin{equation}
    \ell_s=\ell_\text{MAE}+\lambda_\text{RANK} \cdot \ell_\text{RANK}
\end{equation}

\begin{algorithm}[t]
\caption{SimpleVQA-AWP}
\label{alg_simplevqa}
\textbf{Input}: model $\mathbf{f_w}$,batch size $m$,learning rate $\eta_1$, perturbation constraint $\gamma$, perturbation step size $\eta_2$, perturbation steps $K$ \\
\textbf{Output}: model $\mathbf{f_w}$
\begin{algorithmic}[1] 
\REPEAT
\STATE Read mini-batch $B={\mathbf{x}_1,\ldots,\mathbf{x}_m }$ from training set
\FOR{$k=1,\ldots,K$}
\STATE $\boldsymbol{\epsilon} \leftarrow \Pi_\gamma\left(\boldsymbol{\epsilon}+\eta_2 \frac{\nabla_{\boldsymbol{\epsilon}} \frac{1}{m} \sum_{i=1}^m \ell_s\left(\mathbf{f}_{\mathbf{w}+\boldsymbol{\epsilon}}\left(\mathbf{x}_i\right), s_i\right)}{\left\|\nabla_{\boldsymbol{\epsilon}} \frac{1}{m} \sum_{i=1}^m \ell_s\left(\mathbf{f}_{\mathbf{w}+\boldsymbol{\epsilon}}\left(\mathbf{x}_i\right), s_i\right)\right\|}\|\mathbf{w}\|\right)$
\ENDFOR
\STATE
$\mathbf{w} \leftarrow(\mathbf{w}+\boldsymbol{\epsilon})-\eta_1 \nabla_{\mathbf{w}+\boldsymbol{\epsilon}} \frac{1}{m} \sum_{i=1}^m \ell_s\left(\mathbf{f}_{\mathbf{w}+\boldsymbol{\epsilon}}\left(\mathbf{x}_i, s_i\right)\right)-\boldsymbol{\epsilon}$
\UNTIL{training converged}
\end{algorithmic}
\end{algorithm}

\subsection{SAMA-AWP}

Algorithm \ref{alg_sama} outlines the training process for SAMA-AWP. The training loss for SAMA is consistent with that of FAST-VQA, as computed in Equation \ref{eq_loss}.

\begin{algorithm}[t]
\caption{SAMA-AWP}
\label{alg_sama}
\textbf{Input}: model $\mathbf{f_w}$,batch size $m$,learning rate $\eta_1$, perturbation constraint $\gamma$, perturbation step size $\eta_2$, perturbation steps $K$ \\
\textbf{Output}: model $\mathbf{f_w}$
\begin{algorithmic}[1] 
\REPEAT
\STATE Read mini-batch $B={\mathbf{x}_1,\ldots,\mathbf{x}_m }$ from training set
\FOR{$k=1,\ldots,K$}
\STATE $\boldsymbol{\epsilon} \leftarrow \Pi_\gamma\left(\boldsymbol{\epsilon}+\eta_2 \frac{\nabla_{\boldsymbol{\epsilon}} \frac{1}{m} \sum_{i=1}^m \ell\left(\mathbf{f}_{\mathbf{w}+\boldsymbol{\epsilon}}\left(\mathbf{x}_i\right), s_i\right)}{\left\|\nabla_{\boldsymbol{\epsilon}} \frac{1}{m} \sum_{i=1}^m \ell\left(\mathbf{f}_{\mathbf{w}+\boldsymbol{\epsilon}}\left(\mathbf{x}_i\right), s_i\right)\right\|}\|\mathbf{w}\|\right)$
\ENDFOR
\STATE
$\mathbf{w} \leftarrow(\mathbf{w}+\boldsymbol{\epsilon})-\eta_1 \nabla_{\mathbf{w}+\boldsymbol{\epsilon}} \frac{1}{m} \sum_{i=1}^m \ell\left(\mathbf{f}_{\mathbf{w}+\boldsymbol{\epsilon}}\left(\mathbf{x}_i, s_i\right)\right)-\boldsymbol{\epsilon}$
\UNTIL{training converged}
\end{algorithmic}
\end{algorithm}

\begin{table*}[t]
\centering
\begin{tabular}{@{}ccccccccc@{}}
\toprule[1.0pt]
\multirow{2}{*}{AWP} & \multicolumn{2}{c}{$\text{LSVQ}_\text{test}$} & \multicolumn{2}{c}{$\text{LSVQ}_\text{1080p}$} & \multicolumn{2}{c}{KoNViD-1k} & \multicolumn{2}{c}{LIVE-VQC} \\ \cmidrule(l){2-3} \cmidrule(l){4-5} \cmidrule(l){6-7} \cmidrule(l){8-9}
                     & SRCC           & PLCC          & SRCC           & PLCC           & SRCC           & PLCC          & SRCC          & PLCC         \\ \midrule
0                    & 0.876          & 0.877          & 0.779          & 0.814          & 0.859          & 0.855          & 0.823          & 0.844          \\
0.01                 & 0.811          & 0.811          & 0.663          & 0.705          & 0.818          & 0.812          & 0.727          & 0.754          \\
0.001                & \textbf{0.879} & 0.880          & \textbf{0.778} & \textbf{0.814} & \textbf{0.872} & \textbf{0.867} & 0.830          & 0.846          \\
0.0001               & \textbf{0.879} & \textbf{0.881} & \textbf{0.778} & 0.813          & 0.871          & 0.866          & \textbf{0.839} & \textbf{0.851} \\
0.00001              & 0.878          & 0.879          & 0.776          & 0.812          & 0.869          & 0.864          & 0.831          & 0.850          \\ \bottomrule[1.0pt]
\end{tabular}
\caption{Comparison of the generalization performance of models trained on LSVQ using FAST-VQA with AWP of varying perturbation magnitudes across various datasets.}
\label{tab_ablation}
\end{table*}

\section{Implementation Details}
Our basic environment consists of Python 3.11.5, Pytorch 2.1.2, torchvision 0.16.2, and CUDA 12.0. All experiments are conducted using a single NVIDIA A100-SXM4-40GB GPU. Detailed implementations of SimpleVQA-AWP, FAST-VQA-AWP, and SAMA-AWP are provided below.

\subsection{SimpleVQA-AWP}
Following SimpleVQA~\cite{simplevqa}, we employ ResNet50 as the backbone for the spatial feature extraction module and SlowFast R50 for the motion feature extraction model. The weights of ResNet50 are initialized through pretraining on the ImageNet dataset, while the weights of SlowFastR50 are fixed through pretraining on the Kinetics400 dataset. Other weights are randomly initialized. For the spatial feature extraction module, we resize the minimum dimension of the key frame to 520 while maintaining its aspect ratio. During the training phase, input frames are randomly cropped to a resolution of $448\times448$. In the testing phase, the center region is cropped to a resolution of $448\times448$. For the motion feature extraction module, the video resolution is adjusted to $224\times224$ during both the training and testing phases. The proposed model is trained using the Adam optimizer with an initial learning rate of 1e-5 and a batch size of 8. The first frame of each block is selected as the key frame.

\subsection{FAST-VQA-AWP}

Following FAST-VQA~\cite{fastvqa}, we employ VideoSwinTransformer, pretrained on Kinetics 400, as the backbone network. The base learning rate for the backbone is set to 1e-4, while the learning rate for the regression head is set to 1e-3. The model is trained for 30 epochs, with a linear warm-up for the first 2.5 epochs followed by cosine annealing. The batch size is set to 16, and the model is optimized using AdamW. For each training video sample, we sample a $7\times7$ grid, and within each grid, a fragment of size $H\times W\times T=32\times32\times32$ is sampled. Thus, the input dimension is $224\times224\times32$. The embedding patch size is set to $4\times4\times2$. During inference, 128 frames are sampled from the test video and divided into 4 fragments. The final quality score is the average of these fragments.

\subsection{SAMA-AWP}
\subsubsection{VQA}
The implementation of SAMA~\cite{sama} is based on FAST-VQA~\cite{fastvqa}, with the key difference being the replacement of the original fragment sampling with the proposed SAMA. The batch size is adjusted from 16 to 12.

\subsubsection{IQA}
Following SAMA, we employ SwinTransformer, pretrained on ImageNet, as the backbone network. The input size is set to $256\times256$. We sample an $8\times8$ grid, and within each grid, a $32\times32$ patch is sampled. The learning rate for the backbone is set to 1e-4, while the learning rate for the regression head is set to 1e-3. The model is trained for 50 epochs, with a linear warm-up for the first 5 epochs. The batch size is set to 64. The model is optimized using smooth $L_1$ loss and AdamW.

\section{Ablation Studies}

We conducted ablation studies on the results presented in Table \ref{tab_main} to explore the impact of AWP's weight perturbation magnitude on the model's generalization performance on the LSVQ dataset. As shown in Table \ref{tab_ablation}, similar to the results on KoNViD-1k, the best generalization performance is achieved when $\gamma=0.0001$. Increasing the perturbation magnitude beyond this value leads to a decline in generalization performance.


\section{Experiments on Other Datasets}
\begin{figure}[ht]  
\centering
\subfloat[]
{\includegraphics[width=.52\linewidth]{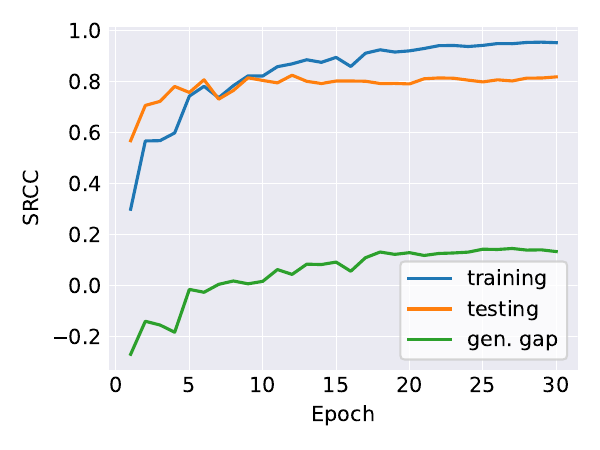}}
\subfloat[]
{\includegraphics[width=.46\linewidth]{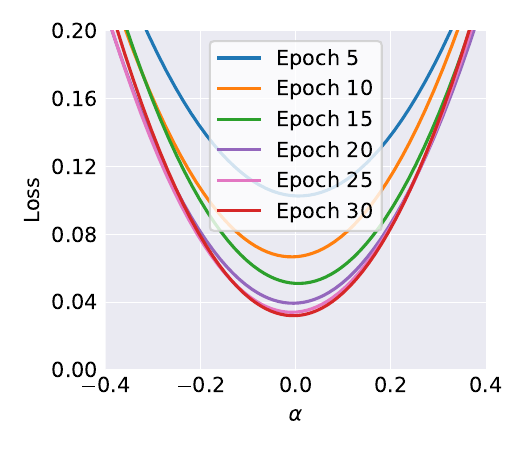}} \\
\caption{The training results of various methods on the YouTubeUGC dataset. (a) illustrates the performance metrics of FAST-VQA on both the training and testing sets across different epochs, as well as the generalization gap. (b) compares the weight loss landscape of FAST-VQA at different epochs.}
\label{fig_epoch_youtubeugc}
\end{figure}

\begin{figure}[ht]
    \centering

    \subfloat[]
    {\includegraphics[width=.48\linewidth]{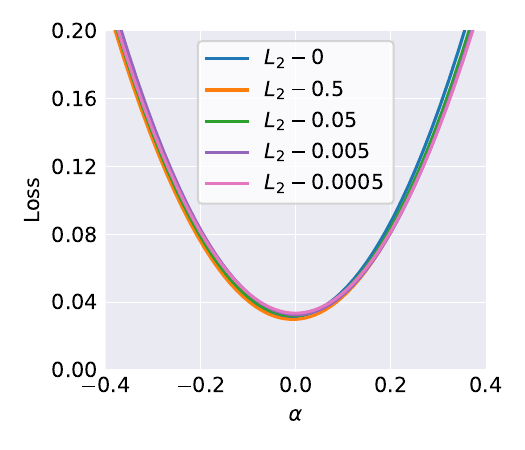}}
    \subfloat[]
    {\includegraphics[width=.48\linewidth]{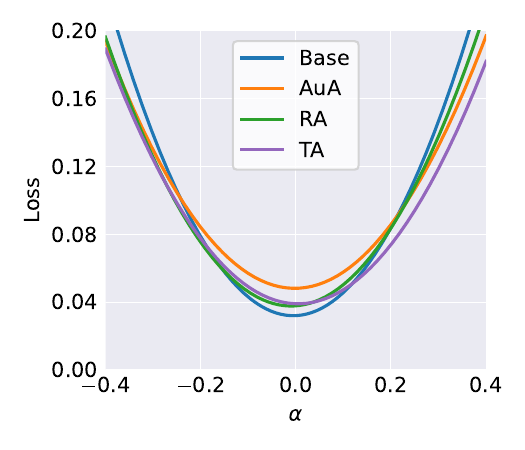}} \\
    \subfloat[]
    {\includegraphics[width=.48\linewidth]{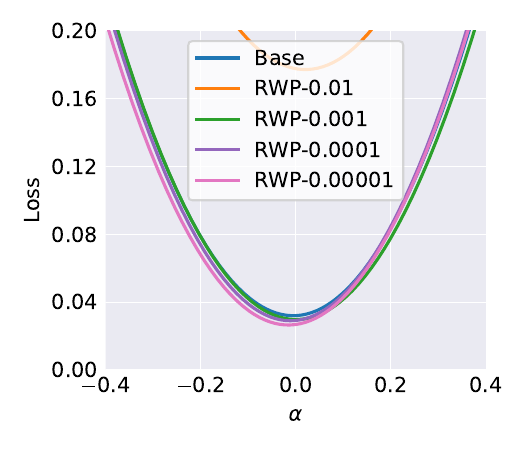}}
    \subfloat[]
    {\includegraphics[width=.48\linewidth]{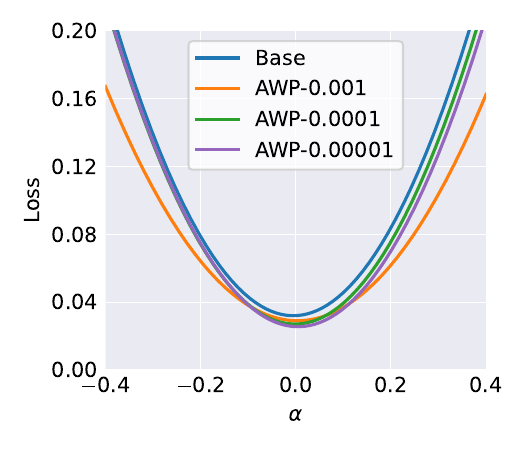}}
    \caption{Weight loss landscape of FAST-VQA on the YouTubeUGC dataset under various modifications. (a) Different $L_2$ regularization parameters $\lambda$, (b) Different data augmentation methods, (c) Different RWP perturbation magnitudes, (d) Different AWP perturbation magnitudes.}
    \label{fig_loss_landscape_youtubeugc}

\end{figure}

In the main text, the exploration of the relationship between the weight loss landscape and the generalization gap, as well as the efforts to reduce the generalization gap, were conducted on the KoNViD-1k dataset. 
In this section, we present the results obtained on the YouTubeUGC~\cite{youtubeugc} dataset, as shown in Tables \ref{tab_l2_youtubeugc}, \ref{tab_aug_youtubeugc}, \ref{tab_rwp_youtubeugc}, \ref{tab_awp_youtubeugc} and Figures \ref{fig_epoch_youtubeugc}, \ref{fig_loss_landscape_youtubeugc}. 
From these results, it is evident that our findings can be generalized to the YouTubeUGC dataset.

\begin{table}[t]
\centering
\setlength{\tabcolsep}{2.2mm}{
\begin{tabular}{@{}ccccccc@{}}
\toprule[1.0pt]
\multirow{2}{*}{Reg.} & \multicolumn{3}{c}{SRCC}                         & \multicolumn{3}{c}{PLCC}                         \\ \cmidrule(l){2-4} \cmidrule(l){5-7}
                      & Train          & Test           & Gap            & Train          & Test           & Gap            \\ \midrule
0                     & \textbf{0.957} & \textbf{0.820} & 0.137          & \textbf{0.959} & 0.828          & 0.131          \\
0.5                   & 0.956          & 0.818          & 0.138          & 0.957          & 0.826          & 0.130          \\
0.05                  & 0.953          & 0.819          & \textbf{0.134} & 0.956          & 0.829          & 0.128          \\
0.005                 & 0.950          & 0.812          & 0.138          & 0.954          & 0.826          & 0.128          \\
0.0005                & 0.948          & 0.814          & 0.135          & 0.951          & \textbf{0.830} & \textbf{0.121} \\ \bottomrule[1.0pt]
\end{tabular}
}
\caption{Performance metrics of FAST-VQA on the YouTubeUGC~\cite{youtubeugc} training and testing sets with different $L_2$ regularization parameters $\lambda$, along with the generalization gap.}
\label{tab_l2_youtubeugc}
\end{table}

\begin{table}[t]
\centering
\setlength{\tabcolsep}{2.3mm}{
\begin{tabular}{@{}ccccccc@{}}
\toprule[1.0pt]
\multirow{2}{*}{Aug.} & \multicolumn{3}{c}{SRCC}                         & \multicolumn{3}{c}{PLCC}                         \\ \cmidrule(l){2-4} \cmidrule(l){5-7} 
                      & Train          & Test           & Gap            & Train          & Test           & Gap            \\ \midrule
base                  & \textbf{0.953} & 0.819          & 0.134          & \textbf{0.956} & 0.829          & 0.128          \\
AuA                   & 0.885          & 0.806          & \textbf{0.080} & 0.883          & 0.827          & \textbf{0.056} \\
RA                    & 0.925          & 0.817          & 0.108          & 0.926          & \textbf{0.833} & 0.093          \\
TA                    & 0.904          & \textbf{0.820} & 0.084          & 0.910          & 0.829          & 0.080          \\ \bottomrule[1.0pt]
\end{tabular}
}
\caption{Performance metrics of FAST-VQA with different data augmentation methods on the YouTubeUGC training and testing sets, along with the generalization gap.}
\label{tab_aug_youtubeugc}
\end{table}

\begin{table}[t]
\centering
\setlength{\tabcolsep}{2.1mm}{
\begin{tabular}{@{}ccccccc@{}}
\toprule[1.0pt]
\multirow{2}{*}{RWP} & \multicolumn{3}{c}{SRCC}                         & \multicolumn{3}{c}{PLCC}                         \\ \cmidrule(l){2-4} \cmidrule(l){5-7} 
                     & Train          & Test           & Gap            & Train          & Test           & Gap            \\ \midrule
0                    & 0.953          & 0.819          & 0.134          & 0.956          & \textbf{0.829} & 0.128          \\
0.01                 & 0.756          & 0.736          & \textbf{0.019} & 0.758          & 0.734          & \textbf{0.024} \\
0.001                & 0.953          & 0.816          & 0.137          & 0.955          & 0.826          & 0.129          \\
0.0001               & \textbf{0.954} & \textbf{0.824} & 0.130          & \textbf{0.957} & \textbf{0.829} & 0.128          \\
0.00001              & \textbf{0.954} & 0.821          & 0.132          & \textbf{0.957} & 0.828          & 0.129          \\ \bottomrule[1.0pt]
\end{tabular}
}
\caption{Performance metrics of FAST-VQA with RWP improvements of varying perturbation magnitudes on the YouTubeUGC training and testing sets, along with the generalization gap.}
\label{tab_rwp_youtubeugc}
\end{table}

\begin{table}[t]
\centering
\setlength{\tabcolsep}{1.7mm}{
\begin{tabular}{@{}ccccccc@{}}
\toprule[1.0pt]
\multirow{2}{*}{AWP} & \multicolumn{3}{c}{SRCC}                          & \multicolumn{3}{c}{PLCC}                          \\ \cmidrule(l){2-4} \cmidrule(l){5-7} 
                     & Train          & Test           & Gap             & Train          & Test           & Gap             \\ \midrule
0                    & 0.953          & 0.819          & 0.134           & 0.956          & 0.829          & 0.128           \\
0.001                & 0.940          & \textbf{0.834} & 0.107           & 0.946          & \textbf{0.842} & 0.104           \\
0.0001               & \textbf{0.956} & 0.817          & 0.139           & \textbf{0.959} & 0.825          & 0.134           \\
0.00001              & 0.953          & 0.827          & 0.126           & 0.956          & 0.834          & 0.122           \\ \bottomrule[1.0pt]
\end{tabular}
}
\caption{Performance metrics of FAST-VQA with AWP improvements of varying perturbation magnitudes on the YouTubeUGC training and testing sets, along with the generalization gap.}
\label{tab_awp_youtubeugc}
\end{table}

\end{document}